\documentclass[11pt,a4paper]{article}
\usepackage[hyperref]{emnlp2018}
\usepackage{times}
\usepackage{latexsym}
\usepackage{graphicx}
\usepackage{amsfonts}
\usepackage{tabularx}

\usepackage{url}

\aclfinalcopy 

\newcolumntype{R}{>{\raggedleft\arraybackslash}X}

\title{Unsupervised Identification of Study Descriptors in Toxicology Research: An Experimental Study}

\author{Drahomira Herrmannova, Steven R. Young, Robert M. Patton, Christopher G. Stahl \\
    Oak Ridge National Laboratory, TN, USA \\
    {\tt \{herrmannovad, youngsr, pattonrm, stahlcg\}@ornl.gov} \\\AND
    Nicole C. Kleinstreuer \\
    NICEATM, NTP, NIEHS, NIH \\
    Research Triangle Park, NC, USA \\
    {\tt nicole.kleinstreuer@nih.gov}\\\AND
    Mary S. Wolfe \\
    NTP, NIEHS, NIH \\
    Research Triangle Park, NC, USA \\
    {\tt wolfe@niehs.nih.gov} \\}

\date{\today}

\begin{document}
\maketitle
\begin{abstract}
Identifying and extracting data elements such as study descriptors in publication full texts is a critical yet manual and labor-intensive step required in a number of tasks. In this paper we address the question of identifying data elements in an unsupervised manner. Specifically, provided a set of criteria describing specific study parameters, such as species, route of administration, and dosing regimen, we develop an unsupervised approach to identify text segments (sentences) relevant to the criteria. A binary classifier trained to identify publications that met the criteria performs better when trained on the candidate sentences than when trained on sentences randomly picked from the text, supporting the intuition that our method is able to accurately identify study descriptors.
\end{abstract}

\section*{Acknowledgments}

Support for this research was provided by a grant from the National Institute of Environmental Health Sciences (AES 16002-001), National Institutes of Health to Oak Ridge National Laboratory.

This research was supported in part by an appointment to the Oak Ridge National Laboratory ASTRO Program, sponsored by the U.S. Department of Energy and administered by the Oak Ridge Institute for Science and Education.

This manuscript has been authored by UT-Battelle, LLC under Contract No. DE-AC05-00OR22725 with the U.S. Department of Energy. The United States Government retains and the publisher, by accepting the article for publication, acknowledges that the United States Government retains a non-exclusive, paid-up, irrevocable, worldwide license to publish or reproduce the published form of this manuscript, or allow others to do so, for United States Government purposes. The Department of Energy will provide public access to these results of federally sponsored research in accordance with the DOE Public Access Plan\footnote{\url{http://energy.gov/downloads/doe-public-access-plan}}.

\section{Introduction}

\begin{figure*}[ht]
    \centering
    \includegraphics[width=\linewidth]{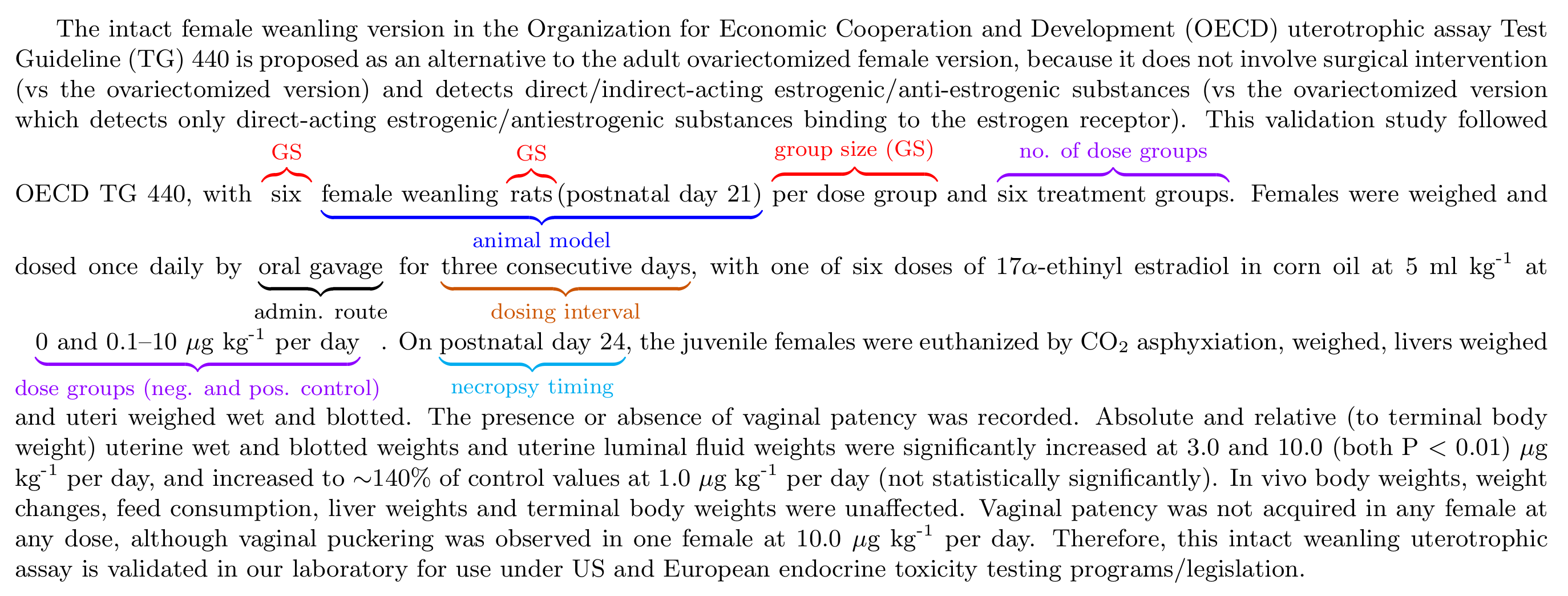}
    \caption{Text excerpt from a reference database of rodent uterotrophic bioassay publications \cite{kleinstreuer_2016_curated}. The text in this example was manually annotated by one of the authors to highlight information relevant to guidelines for performing uterotrophic bioassays set forth by \cite{organisation2007test}.}
    \label{fig:abstract}
\end{figure*}

Extracting data elements such as study descriptors from publication full texts is an essential step in a number of tasks including systematic review preparation \cite{jonnalagadda_2015_automating}, construction of reference databases \cite{kleinstreuer_2016_curated}, and knowledge discovery \cite{smalheiser_2012_literature}. These tasks typically involve domain experts identifying relevant literature pertaining to a specific research question or a topic being investigated, identifying passages in the retrieved articles that discuss the sought after information, and extracting structured data from these passages. The extracted data is then analyzed, for example to assess adherence to existing guidelines \cite{kleinstreuer_2016_curated}. Figure \ref{fig:abstract} shows an example text excerpt with information relevant to a specific task (assessment of adherence to existing guidelines \cite{kleinstreuer_2016_curated}) highlighted.

Extracting the data elements needed in these tasks is a time-consuming and at present a largely manual process which requires domain expertise. For example, in systematic review preparation, information extraction generally constitutes the most time consuming task \cite{tsafnat_2014_systematic}. This situation is made worse by the rapidly expanding body of potentially relevant literature with more than one million papers added into PubMed each year \cite{landhuis2016scientific}. Therefore, data annotation and extraction presents an important challenge for automation.

A typical approach to automated identification of relevant information in biomedical texts is to infer a prediction model from labeled training data -- such a model can then be used to assign predicted labels to new data instances. 
However, obtaining training data for creating such prediction models can be very costly as it involves the step which these models are trying to automate -- manual data extraction. Furthermore, depending on the task at hand, the types of information being extracted may vary significantly. For example, in systematic reviews of randomized controlled trials this information generally includes the \textit{patient} group, the \textit{intervention} being tested, the \textit{comparison}, and the \textit{outcomes} of the study (PICO elements) \cite{tsafnat_2014_systematic}. In toxicology research the extraction may focus on routes of exposure, dose, and necropsy timing \cite{kleinstreuer_2016_curated}. Previous work has largely focused on identifying specific pieces of information such as biomedical events \cite{gonzalez2015recent} or PICO elements \cite{jonnalagadda_2015_automating}. However, depending on the domain and the end goal of the extraction, these may be insufficient to comprehensively describe a given study.

Therefore, in this paper we focus on \textit{unsupervised methods} for identifying text segments (such as sentences or fixed length sequences of words) relevant to the information being extracted. We develop a model that can be used to identify text segments from text documents without labeled data and that only requires the current document itself, rather than an entire training corpus linked to the target document. More specifically, we utilize representation learning methods \cite{mikolov_2013_distributed}, where words or phrases are embedded into the same vector space. This allows us to compute semantic relatedness among text fragments, in particular sentences or text segments in a given document and a short description of the type of information being extracted from the document, by using similarity measures in the feature space. The model has the potential to speed up identification of relevant segments in text and therefore to expedite annotation of domain specific information without reliance on costly labeled data.

We have developed and tested our approach on a reference database of rodent uterotropic bioassays\footnote{\url{https://ntp.niehs.nih.gov/pubhealth/evalatm/test-method-evaluations/endocrine-disruptors/ref-data/edhts.html}} \cite{kleinstreuer_2016_curated} which are labeled according to their adherence to test guidelines set forth in \cite{organisation2007test}. Each study in the database is assigned a label determining whether or not it met each of six main criteria defined by the guidelines; however, the database does not contain sentence-level annotations or any information about where the criteria was mentioned in each publication. Due to the lack of fine-grained annotations, supervised learning methods cannot be easily applied to aid annotating new publications or to annotate related but distinct types of studies. This database therefore presents an ideal use-case for unsupervised approaches.

While our approach doesn't require any labeled data to work, we use the labels available in the dataset to evaluate the approach. We train a binary classification model for identifying publications which satisfied given criteria and show the model performs better when trained on relevant sentences identified by our method than when trained on sentences randomly picked from the text. Furthermore, for three out of the six criteria, a model trained solely on the relevant sentences outperforms a model which utilizes full text. The results of our evaluation support the intuition that semantic relatedness to criteria descriptions can help in identifying text sequences discussing sought after information.

There are two main contributions of this work. We present an unsupervised method that employs representation learning to identify text segments from publication full text which are relevant to/contain specific sought after information (such as number of dose groups). In addition, we explore a new dataset which hasn't been previously used in the field of information extraction.

The remainder of this paper is organized as follows. In the following section we provide more details of the task and the dataset used in this study. In Section \ref{sec:approach} we describe our approach. In Section \ref{sec:evaluation} we evaluate our model and discuss our results. In Section \ref{sec:rel_work} we compare our work to existing approaches. Finally, in Section \ref{sec:conclusion} we provide ideas for further study.

\section{The Task and the Data}
\label{sec:task}

This section provides more details about the specific task and the dataset used in our study which motivated the development of our model.

\subsection{Task Description}

Significant efforts in toxicology research are being devoted towards developing new \textit{in vitro} methods for testing chemicals due to the large number of untested chemicals in use ($>$75,000-80,000 \cite{judson2009toxicity, kleinstreuer_2016_curated}) and the cost and time required by existing \textit{in vivo} methods (2-3 years and millions of dollars per chemical \cite{judson2009toxicity}). To facilitate the development of novel \textit{in vitro} methods and assess the adherence to existing study guidelines, a curated database of high-quality \textit{in vivo} rodent uterotrophic bioassay data extracted from research publications has recently been developed and published \cite{kleinstreuer_2016_curated}. 

The creation of the database followed the study protocol design set forth in \cite{organisation2007test}, which is composed of six minimum criteria (MC, Table \ref{tab:criteria}). An example of information pertaining to the criteria is shown in Figure \ref{fig:abstract}. Only studies which met all six minimum criteria were considered guideline-like (GL) and were included in a follow-up detailed study and the final database \cite{kleinstreuer_2016_curated}. However, of the 670 publications initially considered for inclusion, only 93 ($\sim$14\%) were found to contain studies which met all six MC and could therefore be included in the final database; the remaining 577 publications could not be used in the final reference set. Therefore, significant time and resources could be saved by automating the identification and extraction of the MC.

\begin{table*}[ht!]
\begin{center}
\begin{tabularx}{\textwidth}{lX}
    \bf Criteria name & \bf Description \\ \hline
    MC 1: Animal model & Immature rats, ovariectomized (OVX) adult rats, or OVX adult mice are acceptable (immature mice are not acceptable). OVX animals: OVX should be performed between six and eight weeks of age (allowing at least 14 days post-surgery before dosing for rats and seven days post-surgery for mice). Immature rats: dosing should begin between postnatal day (PND) 18 and PND 21, and be completed by PND 25. \\
    \hline
    MC 2: Group size & Each control group should have a minimum of three animals and each test group should have a minimum of five animals. \\
    \hline
    MC 3: Route of administration & Acceptable routes of administration: oral gavage (p.o.), subcutaneous (s.c.) injection, or intraperitoneal (i.p.) injection. \\
    \hline
    MC 4: Number of dose groups & Minimum of two dose level groups. Must have positive control and negative control. \\
    \hline
    MC 5: Dosing interval & Dosing for a minimum of three consecutive days. Complete by PND 25 in immature animals. \\
    \hline
    MC 6: Necropsy timing & Should be carried out 18-36 hours after the last dose. \\
\end{tabularx}
\end{center}
\caption{\label{tab:criteria} Minimum criteria for guideline-like studies. The descriptions are reprinted here from \cite{kleinstreuer_2016_curated}.}
\end{table*} 

While each study present in the database is assigned a label for each MC determining whether a given MC was met and the pertinent protocol information was manually extracted, there exist no fine-grained text annotations showing the exact location within each publication's full text where a given criteria was met. Therefore, our goal was to develop a model not requiring detailed text annotations that could be used to expedite the annotation of new publications being added into the database and potentially support the development of new reference databases focusing on different domains and sets of guidelines. Due to the lack of detailed annotations, our focus was on identification of potentially relevant text segments.

\subsection{The Dataset}

The version of the database which contains both GL and non-GL studies consists of 670 publications (spanning the years 1938 through 2014) with results from 2,615 uterotrophic bioassays. Specifically, each entry in the database describes one study, and studies are linked to publications using PubMed reference numbers (PMIDs). Each study is assigned seven 0/1 labels -- one for each of the minimum criteria and one for the overall GL/non-GL label. The database also contains more detailed subcategories for each label (for example ``species'' label for MC 1) which were not used in this study. The publication PDFs were provided to us by the database creators. We have used the Grobid\footnote{\url{https://github.com/kermitt2/grobid}} library to convert the PDF files into structured text. After removing documents with missing PDF files and documents which were not converted successfully, we were left with 624 full text documents.

Each publication contains on average 3.7 studies (separate bioassays), 194 publications contain a single study, while the rest contain two or more studies (with 82 being the most bioassays per publication). The following excerpt shows an example sentence mentioning multiple bioassays (with different study protocols):


\begin{quote}
    \emph{With the exception of the first study (experiment 1), which had group sizes of~12, all other studies had group sizes of~8.}
\end{quote}

For this experiment we did not distinguish between publications describing a single or multiple studies. Instead, our focus was on retrieving all text segments (which may be related to multiple studies) relevant to each of the criteria. For each MC, if a document contained multiple studies with different labels, we discarded that document from our analysis of that criteria; if a document contained multiple studies with the same label, we simply combine all those labels into a single label.
Table \ref{tab:dataset} shows the final size of the dataset.

\begin{table}[t!]
\begin{center}
\begin{tabularx}{\linewidth}{lXXXX}
    \bf Criteria & \bf 0 & \bf 1 & \bf Total & \bf \% of 1 \\ \hline \hline
    MC 1 & 414 & 175 & 589 & 29.71 \\ \hline
    MC 2 & 35 & 577 & 612 &	94.28 \\ \hline
    MC 3 & 70 & 536 & 606 & 88.45 \\ \hline
    MC 4 & 309 & 206 & 515 & 40.00 \\ \hline
    MC 5 & 96 & 490 & 586 & 83.62 \\ \hline
    MC 6 & 228 & 340 & 568 & 59.86 \\ \hline
    GL & 522 & 72 & 594 & 12.12 \\
\end{tabularx}
\end{center}
\caption{\label{tab:dataset} Label statistics. Column \textit{0} shows number of publications per MC which did not meet the criteria and column \textit{1} shows number of publications which met the criteria. The last column in the table shows proportion of positive (i.e. criteria met) labels.}
\end{table}

\section{Approach}
\label{sec:approach}

In this section we describe the method we have used for retrieving text segments related to the criteria described in the previous section. The intuition is based off question answering systems. We treat the criteria descriptions (Table \ref{tab:criteria}) as the question and the text segments within the publication that discusses the criteria as the answer. Given a full text publication, the goal is to find the text segments most likely to contain the answer.

We represent the criteria descriptions and text segments extracted from the documents as vectors of features, and utilize relatedness measures to retrieve text segments most similar to the descriptions. A similar step is typically performed by most question answering (QA) systems -- in QA systems both the input documents and the question are represented as a sequence of embedding vectors and a retrieval system then compares the document and question representations to retrieve text segments most likely containing the answer \cite{mishra2016survey}.

To account for the variations in language that can be used to describe the criteria, we represent words as vectors generated using Word2Vec \cite{mikolov_2013_distributed}. The following two excerpts show two different ways MC 6 was described in text:

\begin{quote}
    \emph{Animals were killed 24 h after being injected and their uteri were removed and weighed.}
\end{quote}

\begin{quote}
    \emph{All animals were euthanized by exposure to ethyl ether 24 h after the final treatment.}
\end{quote}

\begin{figure}[ht!]
    \centering
    \includegraphics[width=\linewidth]{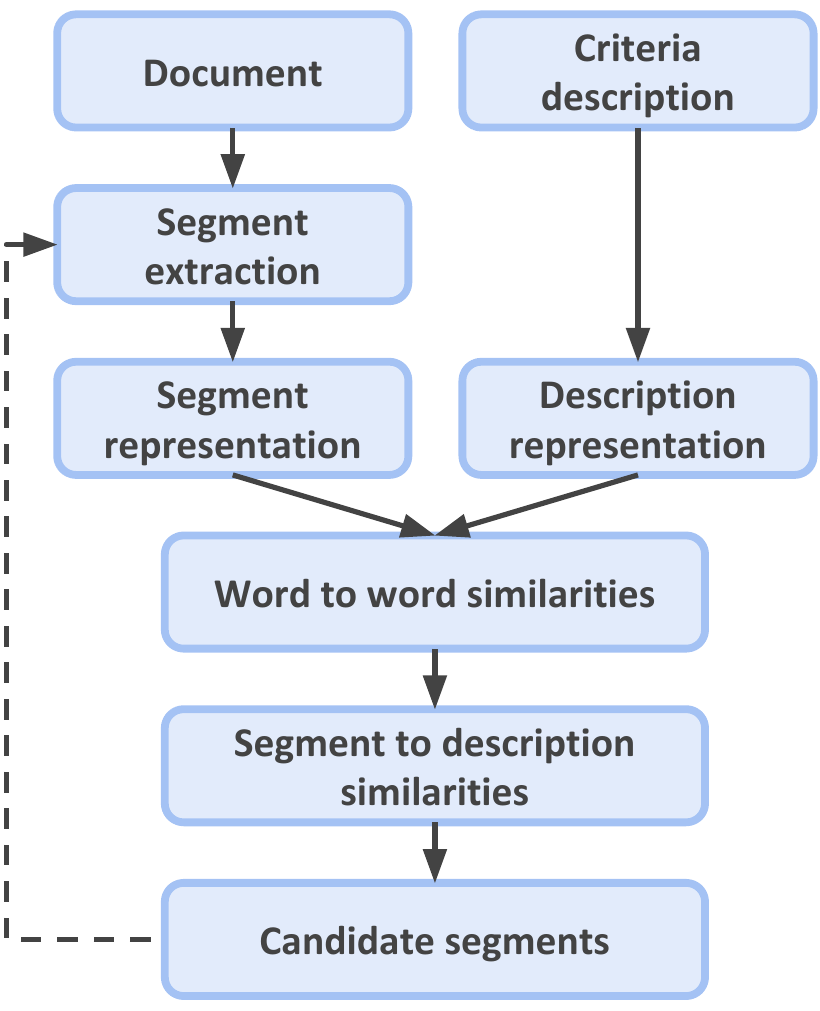}
    \caption{High level overview of our approach. The dotted line represents an optional step of finding smaller sub-segments within the candidate segments. For example, in our case, we first retrieve the most similar sentences and in the second step find most similar continuous 5-grams found withing those sentences.}
    \label{fig:sim_calc}
\end{figure}

We hypothesize that the use of word embedding features will allow us to detect relevant words which are not present in the criteria descriptions. \cite{mikolov2013linguistic} have shown that an important feature of Word2Vec embeddings is that similar words will have similar vectors because they appear in similar contexts. We utilize this feature to calculate similarity between the criteria descriptions and text segments (such as sentences) extracted from each document. A high-level overview of our approach is shown in Figure~\ref{fig:sim_calc}.

We use the following method to retrieve the most relevant text segments:

\textbf{Segment extraction:} First, we break each document down into shorter sequences such as sentences or word sequences of fixed length. While the first option (sentences) results in text which is easier to process, it has the disadvantage of resulting in sequences of varying length which may affect the resulting similarity value. However, for simplicity, in this study we utilize the sentence version.

\textbf{Segment/description representation:} We represent each sequence and the input description as a sequence of vector representations. For this study we have utilized Word2Vec embeddings \cite{mikolov_2013_distributed} trained using the Gensim library on our corpus of 624 full text publications.

\textbf{Word to word similarities:} Next we calculate similarity between each word vector from each sequence $s_i$ and each word vector from the input description $d$ using \textit{cosine similarity}. The output of this step is a similarity matrix $\mathbf{S}_i \in \mathbb{R}^{N_i \times M_d}$ for each sequence $s_i$, where $N_i$ is the number of unique words in the sequence and $M_d$ is the number of unique words in the description $d$.

\textbf{Segment to description similarities:} To obtain a similarity value representing the relatedness of each sequence to the input description we first convert each input matrix $\mathbf{S}_i$ into a vector $v_i \in \mathbb{R}^{N_i}$ by choosing the maximum similarity value for each word in the sequence, that is $v_i = \textrm{max}_{rows}(\mathbf{S}_i)$. Each sequence is then assigned a similarity value $r_i \in \mathbb{R}$ which is calculated as $r_i = \textrm{avg}(v_i)$. In the future we are planning to experiment with different ways of calculating relatedness of the sequences to the descriptions, such as with computing similarity of embeddings created from the text fragments using approaches like Doc2Vec \cite{le_2014_distributed}. In this study, after finding the top sentences, we further break each sentence down into continuous n-grams to find the specific part of the sentence discussing the MC. We repeat the same process described above to calculate the relatedness of each n-gram to the description.

\textbf{Candidate segments:} For each document we select the top $k$ text segments (sentences in the first step and 5-grams in the second step) most similar to the description.

\subsection{Example Results}

\begin{figure*}[ht]
    \centering
    \includegraphics[width=\linewidth]{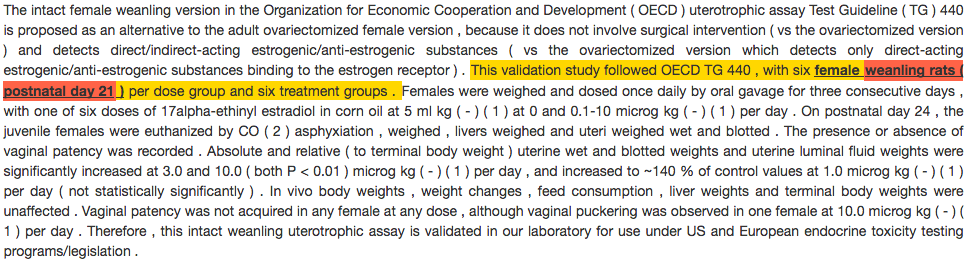}
    \caption{Annotations generated using our method for the abstract from Figure \ref{fig:abstract}. The sentence which was found to be the most similar to the description for ``MC 1: Animal model'' is highlighted in yellow and the most similar sequence of words within that sentence is highlighted in red. The text we are looking for is highlighted with bold underlined text. For this example we ran our method on the abstract of the target document rather than the full text and highlighted only the single most similar sentence.}
    \label{fig:mc1}
\end{figure*}

\begin{figure*}[ht]
    \centering
    \includegraphics[width=\linewidth]{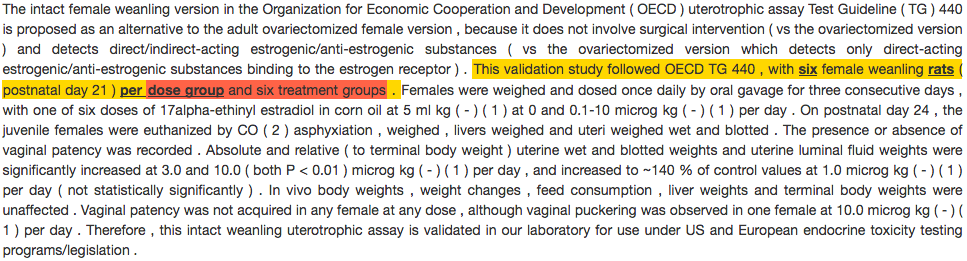}
    \caption{Annotations generated using our method for ``MC 2: Group size''. The highlighting used is the same as in Figure \ref{fig:mc1}.}
    \label{fig:mc2}
\end{figure*}

\begin{figure*}[ht]
    \centering
    \includegraphics[width=\linewidth]{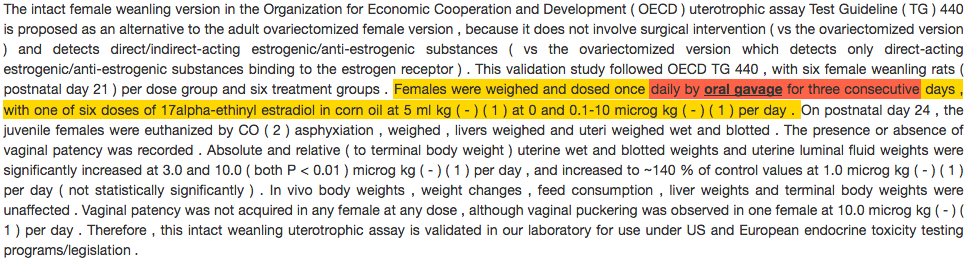}
    \caption{Annotations generated using our method for ``MC 3: Route of administration''. The highlighting used is the same as in Figure \ref{fig:mc1}.}
    \label{fig:mc3}
\end{figure*}

Figures \ref{fig:mc1}, \ref{fig:mc2}, and \ref{fig:mc3} show example annotations generated using our method for the first three criteria. For this example we ran our method on the abstract of the target document rather than the full text and highlighted only the single most similar sentence. The abstract used to produce these figures is the same as the abstract shown in Figure \ref{fig:abstract}. In all three figures, the lighter yellow color highlights the sentence which was found to be the most similar to a given MC description, the darker red color shows the top 5-gram found within the top sentence, and the bold underlined text is the text we are looking for (the correct answer). Annotations generated for the remaining three criteria are shown in Appendix \ref{sec:supplemental}.

Due to space limitations, Figures \ref{fig:mc1}, \ref{fig:mc2}, and \ref{fig:mc3} show results generated on abstracts rather than on full text; however, we have observed similarly accurate results when we applied our method to full text. The only difference between the abstracts and the full text version is how many top sentences we retrieved. When working with abstracts only, we observed that if the criteria was discussed in the abstract, it was generally sufficient to retrieve the single most similar sentence. However, as the criteria may be mentioned in multiple places within the document, when working with full text documents we have retrieved and analyzed the top \textit{k} sentences instead of just a single sentence. In this case we have typically found the correct sentence/sentences among the top 5 sentences. We have also observed that the similar sentences which don't discuss the criteria directly (i.e. the ``incorrect'' sentences) typically discuss related topics. For example, consider the following three sentences:

\begin{quote}
    \emph{After weaning on pnd 21, the dams were euthanized by CO2 asphyxiation and the juvenile females were individually housed.}
\end{quote}    
\begin{quote}
    \emph{Six CD(SD) rat dams, each with reconstituted litters of six female pups, were received from Charles River Laboratories (Raleigh, NC, USA) on offspring postnatal day (pnd) 16.}
\end{quote}
\begin{quote}
    \emph{This validation study followed OECD TG 440, with six female \underline{weanling rats (postnatal day 21)} per dose group and six treatment groups.}
\end{quote}

These three sentences were extracted from the abstract and the full text of a single document (document 20981862, the abstract of which is shown in Figures \ref{fig:abstract} and \ref{fig:mc1}-\ref{fig:mc6}). These three sentences were retrieved as the most similar to MC 1, with similarity scores of 70.61, 65.31, and 63.69, respectively. The third sentence contains the ``answer'' to MC 1 (underlined). However, it can be seen the top two sentences also discuss the animals used in the study (more specifically, the sentences discuss the animals' housing and their origin). 

\section{Evaluation}
\label{sec:evaluation}

The goal of this experiment was to explore empirically whether our approach truly identifies mentions of the minimum criteria in text. As we did not have any fine-grained annotations that could be used to directly evaluate whether our model identifies the correct sequences, we have used a different methodology. We have utilized the existing 0/1 labels which were available in the database (these were discussed in Section \ref{sec:task}) to train one binary classifier for each MC. The task of each of the classifiers is to determine whether a publication met the given criteria or not. We have then compared a baseline classifier trained on all full text with three other models:

\begin{itemize}
    \item A model which, instead of all full text, utilized only the top \textit{k} sentences most similar to the given MC. The top \textit{k} sentences were identified using our model introduced in the previous section.
    \item A model which utilized only the \textit{k} least similar sentences.
    \item A model which utilized only \textit{k} random sentences (but none of the top or bottom \textit{k} sentences -- the sentences were chosen at random from the interval $(k, n-k)$ where $n$ is the number of sentences in the document and where sentences are sorted from the most similar to the least similar).
\end{itemize}

The only difference between the four models is which sentences from each document are passed to the classifier for training and testing. The intuition is that a classifier utilizing the correct sentences should outperform both other models.

To avoid selecting the same sentences across the three models we removed documents which contained less than $3 * k$ sentences (Table \ref{tab:results}, row \textit{Number of documents} shows how many documents satisfied this condition). In all of the experiments presented in this section, the publication full text was tokenized, lower-cased, stemmed, and stop words were removed. All models used a Bernoulli Na{\"i}ve Bayes classifier (scikit-learn implementation which used a uniform class prior) trained on binary occurrence matrices created using 1-3-grams extracted from the publications, with n-grams appearing in only one document removed. The complete results obtained from leave-one-out cross validation are shown in Table \ref{tab:results}. In all cases we report classification accuracy. In the case of the \textit{random-k sentences} model the accuracy was averaged over 10 runs of the model. 

\begin{table*}[ht!]
\begin{center}
\begin{tabularx}{\linewidth}{lRRRRRR}
    \bf Approach & \bf MC1 & \bf MC2 & \bf MC3 & \bf MC4 & \bf MC5 & \bf MC6 \\ 
    \hline
    \hline
    Baseline 1: Most frequent label & 70.35 & 94.43 & 88.74 & 59.48 & 84.30 & 60.44 \\ \hline
    Baseline 2: All full text & 78.25 & 92.06 & 89.59 & 67.94 & 84.83 & 74.05 \\ \hline \hline
    Top-k sentence & 76.84 & 91.55 & 87.71 & 68.35 & 88.54 & 74.23 \\ \hline
    Bottom-k sentences & 70.00 & 91.39 & 88.23 & 63.10 & 80.60 & 63.70 \\ \hline
    Random-k sentences & 73.26 & 93.72 & 88.43 & 65.65 & 85.29 & 68.28 \\
    \hline 
    \hline
    Number of documents & 570 & 592 & 586 & 496 & 567 & 551 \\
    Number of pos. labels & 169 & 559 & 520 & 201 & 478 & 333 \\
\end{tabularx}
\end{center}
\caption{\label{tab:results} Evaluation results.}
\end{table*}

We compare the results to two baselines: (1) a baseline obtained by classifying all documents as belonging to the majority class (\textit{baseline 1} in Table \ref{tab:results}) and (2) a baseline obtained using the same setup (features and classification algorithm) as in the case of the \textit{top-/random-/bottom-k sentences} models but which utilized all full text instead of selected sentences extracted from the text only (\textit{baseline 2} in Table \ref{tab:results}).

\subsection{Results analysis}

Table \ref{tab:results} shows that for four out of the six criteria (MC 1, MC 4, MC 5, and MC 6) the \textit{top-k sentences} model outperforms \textit{baseline 1} as well the \textit{bottom-k} and the \textit{random-k sentences} models by a significant margin. Furthermore, for three of the six criteria (MC 4, MC 5, and MC 6) the \textit{top-k sentences} model also outperforms the \textit{baseline 2} model (model which utilized all full text). This seems to confirm our hypothesis that semantic relatedness of sentences to the criteria descriptions helps in identifying sentences discussing the criteria. These seems to be the case especially given that for three of the six criteria the \textit{top-k sentences} model outperforms the model which utilizes all full text (\textit{baseline 2}) despite being given less information to learn from (selected sentences only in the case of the \textit{top-k sentences} model vs. all full text in the case of the \textit{baseline 2} model).

For two of the criteria (MC 2 and MC 3) this is not the case and the \textit{top-k sentences} model performs worse than both other models in the case of MC 3 and worse than the \textit{random-k} model in the case of MC 2. One possible explanation for this is class imbalance. In the case of MC 2, only 33 out of 592 publications (5.57\%) represent negative examples (Table \ref{tab:results}). As the \textit{top-k sentences} model picks only sentences closely related to MC 2, it is possible that due to the class imbalance the top sentences don't contain enough negative examples to learn from. On the other hand, the \textit{bottom-k} and \textit{random-k sentences} models may select text not necessarily related to the criteria but potentially containing linguistic patterns which the model learns to associate with the criteria; for example, certain chemicals may require the use of a certain study protocol which may not be aligned with the MC and the model may key in on the appearance of these chemicals in text rather than the appearance of MC indicators. The situation is similar in the case of MC 3. We would like to emphasize that the goal of this experiment was not to achieve state-of-the-art results but to investigate empirically the viability of utilizing semantic relatedness of text segments to criteria descriptions for identifying relevant segments.

\section{Related Work}
\label{sec:rel_work}

In this section we present studies most similar to our work. We focus on unsupervised methods for information extraction from biomedical texts. 

Many methods for biomedical data annotation and extraction exist which utilize labeled data and supervised learning approaches (\cite{liu2016learning} and \cite{gonzalez2015recent} provided a good overview of a number of these methods); however, unsupervised approaches in this area are much scarcer. One such approach has been introduced by \cite{zhang2013unsupervised}, who have proposed a model for unsupervised Named Entity Recognition. Similar to our approach, their model is based on calculating the similarity between vector representations of candidate phrases and existing entities. However, their vector representations are created using a combination of TF-IDF weights and word context information, and their method relies on a terminology. More recently, \cite{chen2018word2vec} have utilized Word2Vec and Doc2Vec embeddings for unsupervised sentiment classification in medical discharge summaries.

A number of previous studies have focused on unsupervised extraction of relations such as protein-protein interactions (PPI) from biomedical texts. For example, \cite{quan2014unsupervised} have utilized several techniques, namely kernel-based pattern clustering and dependency parsing, to extract PPI from biomedical texts. \cite{alicante2016unsupervised} have introduced a system for unsupervised extraction of entities and relations between these entities from clinical texts written in Italian, which utilized a thesaurus for extraction of entities and clustering methods for relation extraction. \cite{rink2011generative} also used clinical texts and proposed a generative model for unsupervised relation extraction. Another approach focusing on relation extraction has been proposed by \cite{madkour2007bionoculars}. Their approach is based on constructing a graph which is used to construct domain-independent patterns for extracting protein-protein interactions.

A similar but distinct approach to unsupervised extraction is distant supervision. Similarly as unsupervised extraction methods, distant supervision methods don't require any labeled data, but make use of weakly labeled data, such as data extracted from a knowledge base. Distant supervision has been applied to relation extraction \cite{liu2014relation}, extraction of gene interactions \cite{mallory2015large}, PPI extraction \cite{thomas2012weakly,bobic2012improving}, and identification of PICO elements \cite{wallace2016extracting}. The advantage of our approach compared to the distantly supervised methods is that it does not require any underlying knowledge base or a similar source of data.


\section{Conclusions and Future Work}
\label{sec:conclusion}

In this paper we presented a method for unsupervised identification of text segments relevant to specific sought after information being extracted from scientific documents. Our method is entirely unsupervised and only requires the current document itself and the input descriptions instead of corpus linked to this document. The method utilizes short descriptions of the information being extracted from the documents and the ability of word embeddings to capture word context. Consequently, it is domain independent and can potentially be applied to another set of documents and criteria with minimal effort. We have used the method on a corpus of toxicology documents and a set of guideline protocol criteria needed to be extracted from the documents. We have shown the identified text segments are very accurate. Furthermore, a binary classifier trained to identify publications that met the criteria performed better when trained on the candidate sentences than when trained on sentences randomly picked from the text, supporting our intuition that our method is able to accurately identify relevant text segments from full text documents. 

There are a number of things we plan on investigating next. In our initial experiment we have utilized criteria descriptions which were not designed to be used by our model. One possible improvement of our method could be replacing the current descriptions with example sentences taken from the documents containing the sought after information. We also plan on testing our method on an annotated dataset, for example using existing annotated PICO element datasets \cite{boudin2010combining}.



\bibliography{emnlp2018}
\bibliographystyle{acl_natbib_nourl}

\clearpage

\appendix
\section{Supplemental Material}
\label{sec:supplemental}

\begin{figure*}[h]
    \centering
    \includegraphics[width=\linewidth]{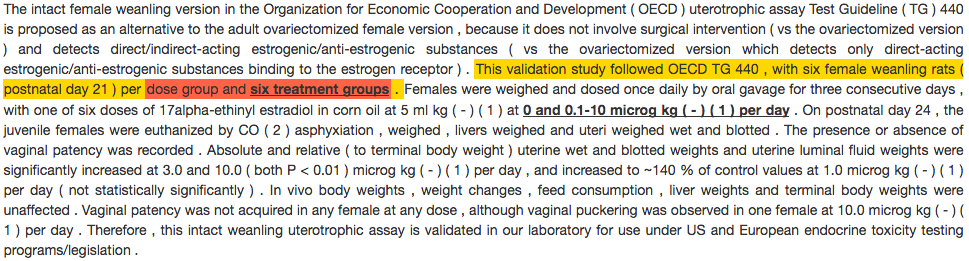}
    \caption{Annotations generated using our method for abstract from Figure \ref{fig:abstract}. The sentence which was found to be the most similar to the description for ``MC 4: Number of dose groups'' is highlighted in yellow and the most similar sequence of words within that sentence is highlighted in red. The text we are looking for is highlighted with bold underlined text. For this example we ran our method on the abstract of the target document rather than the full text and highlighted only the single most similar sentence.}
    \label{fig:mc4}
\end{figure*}

\begin{figure*}[h]
    \centering
    \includegraphics[width=\linewidth]{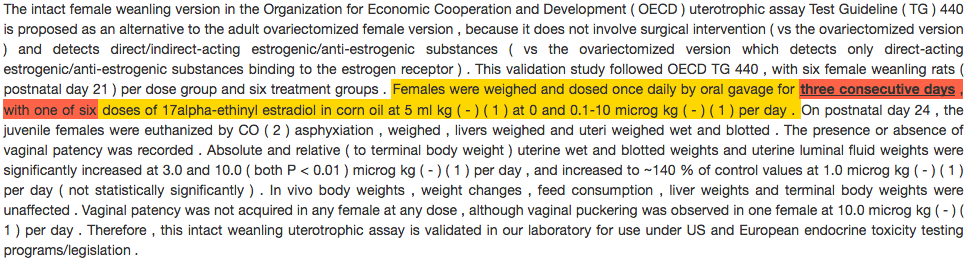}
    \caption{Annotations generated using our method for ``MC 5: Dosing interval''.}
    \label{fig:mc5}
\end{figure*}

\begin{figure*}[h]
    \centering
    \includegraphics[width=\linewidth]{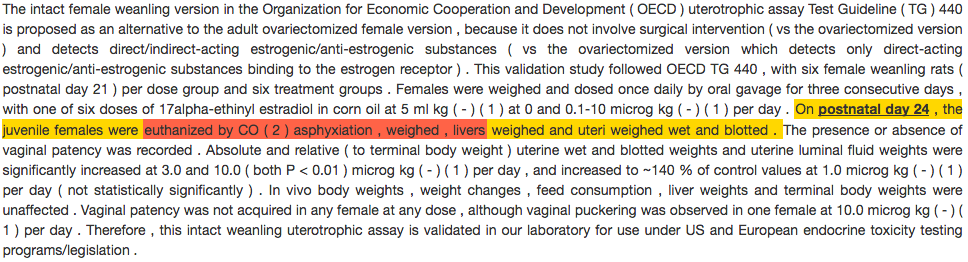}
    \caption{Annotations generated using our method for ``MC 6: Necropsy timing''.}
    \label{fig:mc6}
\end{figure*}

This section provides additional details and results. Figures \ref{fig:mc4}, \ref{fig:mc5}, and \ref{fig:mc6} show example annotations generated for criteria MC 4, MC 5, and MC 6.

\end{document}